\documentclass[letterpaper, 10 pt, conference]{ieeeconf}  

\IEEEoverridecommandlockouts                              

\usepackage{graphicx}
\usepackage{subfigure}
\usepackage{amsmath} 
\usepackage{amssymb}  
\usepackage{bm}

\title{\LARGE \bf
Mind Your Manners! A Dataset and A Continual Learning Approach for Assessing Social Appropriateness of Robot Actions\\
}

\author{Jonas Tjomsland$^{1}$, Sinan Kalkan$^{2}$ and Hatice Gunes$^{1}$
\thanks{$^{1}$J. Tjomsland and H. Gunes are with the Department of Computer Science and Technology, University of Cambridge, United Kingdom.
        {\tt\small \{jonas.tjomsland, hatice.gunes\}@cl.cam.ac.uk}}%
\thanks{$^{2}$S. Kalkan is with the Department of Computer Engineering, Middle East Technical University, Ankara, Turkey, and currently a visiting researcher at the Department of Computer Science and Technology, University of Cambridge, United Kingdom.
        {\tt\small skalkan@metu.edu.tr}}%
\thanks{J. Tjomsland and H. Gunes' work has been partially supported by the EPSRC under grant ref: EP/R030782/1. S. Kalkan is supported by Scientific and Technological Research Council of Turkey (T\"UB\.ITAK) through BIDEB 2219 International Postdoctoral Research Scholarship Program.}
}

\begin{document}

\maketitle
\thispagestyle{empty}
\pagestyle{empty}

\begin{abstract}
To date, endowing robots with an ability to assess social appropriateness of their actions has not been possible. This has been mainly due to  (i) the lack of relevant and labelled data, and (ii) the lack of formulations of this as a lifelong learning problem. In this paper, we address these two issues. We first introduce the Socially Appropriate Domestic Robot Actions dataset (MANNERS-DB), which contains appropriateness labels of robot actions annotated by humans. To be able to control but vary the configurations of the scenes and the social settings, MANNERS-DB has been created utilising a simulation environment by uniformly sampling relevant contextual attributes. Secondly, we train and evaluate a baseline Bayesian Neural Network (BNN) that estimates social appropriateness of actions in the MANNERS-DB. Finally, we formulate learning social appropriateness of actions as a continual learning problem using the uncertainty of the BNN parameters. The experimental results show that the social appropriateness of robot actions can be predicted with a satisfactory level of precision. Our work takes robots one step closer to a human-like understanding of (social) appropriateness of actions, with respect to the social context they operate in. To facilitate reproducibility and further progress in this area, the MANNERS-DB, the trained models and the relevant code will be made publicly available.
\end{abstract}


\section{INTRODUCTION}
Similarly to humans, social robots, which are expected to inhabit highly challenging environments populated with complex objects, articulated tools, and complicated social settings involving humans, animals and other robots, should be able to assess whether an action is socially appropriate in a given context. The social robotics community has studied related problems such as socially appropriate navigation \cite{gomez2013social}, recognition of human intent \cite{losey2018review}, engagement \cite{SalamEtAl2017}, facial expressions and personality \cite{GunesEtAl2019}. However, determining whether generic robot actions are appropriate or not in a given social context is a relatively less explored area of research.

\begin{figure}[h]
\centering
\includegraphics[width=0.49\textwidth]{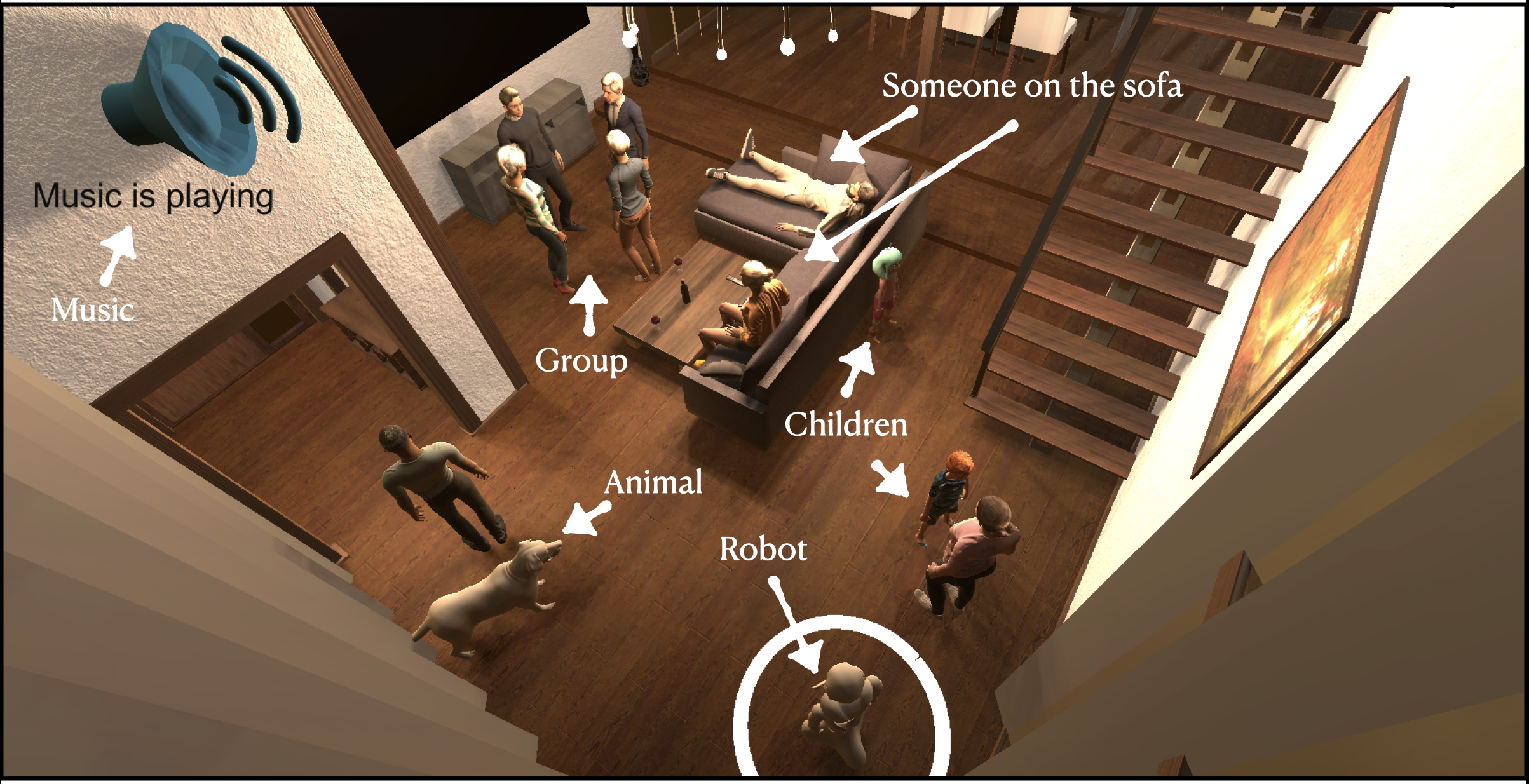}
\caption{An example scene of the simulated living room environment.\label{fig:first_scene}}
\end{figure}

Our work takes robots one step closer to a human-like understanding of (social) appropriateness of actions, with respect to the social context they operate in. To this end, we first introduce a dataset called MANNERS-DB that constitutes simulated  robot actions in visual domestic scenes of different social configurations (see an example in Fig. \ref{fig:first_scene}). The robot actions in each scene have been annotated by humans with social appropriateness levels. Moreover, we train and evaluate a baseline Bayesian Neural Network that estimates social appropriateness of actions on the MANNERS-DB. Finally, we formulate learning social appropriateness of actions as a continual learning problem and propose a Bayesian continual learning model that can incrementally learn social appropriateness of new actions.

\section{Related Work}

\subsection{Social Appropriateness and HRI}

The well-known survey paper on social signal processing by Vinciarelli et al. \cite{VINCIARELLI20091743} provides a compilation of the relevant cues associated to the most important social behaviours, including behavioural cues related to posture and usage of space.
Kendon in \cite{Kendon2009} proposed the \emph{Facing-formation system of spatial organisation} where \emph{F-formations} refer to the spatial patterns formed when people interact face-to-face. 
According to this framework, the space that an individual directs their attention to is called a transactional segment.
When two or more people's transactional segments overlap during an interaction, an F-formation with different configurations is formed (L-arrangement, face-to-face, side-by-side, semicircular, and rectangular arrangements). This framework has been widely adopted for automatic analysis of free-standing group interactions. 
%
%
%
When it comes to assessing how we use the space and the environment around us in social interactions, Hall \cite{hall1968proxemics} identified four concentric zones around a person, namely the intimate, the casual-personal, the socio-consultive and the public zone. He argued that the preferred interpersonal distance between the parties interacting is determined by the social relationship between them. The  \textit{intimate zone} is reserved for our closest relations which we embrace and physically touch. The \textit{casual-personal zone}, is where we interact with friends and wider family. The \textit{socio-consultive zone} is where acquaintances and etc. are let in. And lastly, the \textit{public zone} is where strangers and impersonal interactions often occur.

In the field of human-robot interaction, studies have shown that robots are treated differently than humans with respect to appropriate interpersonal distance and invasion of personal space. Evidence suggests that, when introduced to a robot, people prefer it to be positioned in what Hall \cite{hall1968proxemics} defines as the \textit{social zone} and only after first interactions they would feel comfortable allowing it into their \textit{personal zone} \cite{huttenrauch2006investigating, walters2009empirical}. Studies also show that these preferences change as people get used to the robot over time \cite{koay2007living}, and that these preferences are also dependent on robot appearance \cite{walters2008avoiding}. We note that majority of the existing work on robot behaviour toward and around people have focused on socially aware motion planning \cite{triebel2016spencer} or approach \cite{walters2007robotic}. Researchers have also examined how and when to engage humans appropriately in HRI situations, based on sensory inputs indicating location, pose, and movement of humans in the environment \cite{michalowski2006spatial}. To the best of our knowledge, the perception and the machine-learning based recognition of social appropriateness of domestic robot actions has not been investigated. 

\subsection{Continual Learning}
Humans excel at continuously learning new skills and new knowledge with new experiences. This has inspired a new problem in machine learning, coined as continual learning (CL), also as lifelong learning or incremental learning  \cite{thrun1995lifelong, chen2016lifelong,ring1994continual,lesort2020continual}, that studies developing methods that can continuously learn with each new experience or task. Unlike conventional machine learning models, CL models do not assume the data, its distribution or the tasks to be fixed. 

An important challenge in CL is to be able to retain the previously acquired knowledge while learning new ones. Called as the catastrophic forgetting problem \cite{french1999catastrophic,mccloskey1989catastrophic}, unless measures are taken, learning from new experience tends to overwrite the previously learned associations. 

Over the years, many strategies have been devised against catastrophic forgetting (see \cite{thrun1995lifelong,Parisi2018b} for reviews). These include e.g. regularizing the destructive supervision signals, rehearsing previously obtained experiences, modifying the architecture by increasing new neurons or layers, or employing neuro-inspired approaches such as using an episodic memory with consolidation of novel experiences to a long-term memory. In this paper, we use a method that regularizes updates to parameters by looking at their uncertainties, following the approach in \cite{ebrahimi2019uncertainty}.
\section{THE MANNERS DATASET and ITS LABELS}

Since it is difficult to create a real environment with simultaneously controlled and varied social configurations and attributes, we developed a simulation environment to generate static scenes with various social configurations and attributes. The scenes were then labeled by independent observers via a crowdsourcing platform.

\subsection{Dataset Generation} 
\textbf{The Simulation Environment}. The environment was developed in Unity3D simulation software \cite{unity}. With Unity, we could generate a living room scene with various social configurations and attributes, involving groups of people, children and animals in different orientations and positions in space, music playing, robot facing the people etc. See an example scene with these aspects illustrated in Fig. \ref{fig:first_scene}. The living room in which all the scenes are generated is part of a Unity Asset package from Craft Studios \cite{apartment}. All avatars used to represent either people or animals are taken from Adobe's Mixamo software \cite{avatar}. Avatars are spawned into the living room scene as Unity Gameobjects, following a script written in the Unity compatible C\# programming language.

\textbf{Scene Generation}.
1000 scenes were generated by uniformly sampling the factors listed in Table \ref{table:feat}, which include the number of people, the number of groups with people, animals, their locations and orientations etc. Specific attention was paid to the uniform sampling of positions and orientations to ensure that the dataset contains a wide spectrum of proxemics \cite{hall1968proxemics,Kendon2009} configurations.

\textbf{Robot actions}. We specifically consider the social appropriateness of the actions listed in Table \ref{table:actions}. In total, 16 robot actions are investigated \-- all actions except \textit{Cleaning (Picking up stuff)} and \textit{Starting a conversation} are investigated in two sets, based on whether they are executed in a region (within a circle surrounding the robot) or in a direction pointed by an arrow.

\begin{table}[ht]
\caption{The robot actions investigated in each scene}
\centering
\begin{tabular}{|l|l|} \hline
\emph{Actions within a circle} & \emph{Actions along an arrow} \\\hline\hline
Vacuum cleaning & Vacuum cleaning \\ 
\hline
Mopping the floor & Mopping the floor \\ 
\hline
Carry warm food & Carry warm food \\ 
\hline
Carry cold food & Carry cold food \\ 
\hline
Carry drinks & Carry drinks \\ 
\hline
Carry small objects (plates, toys) & Carry small objects (plates, toys) \\ 
\hline
Carry big objects (tables, chairs) & Carry big objects (tables, chairs) \\ 
\hline
Cleaning (Picking up stuff) & Starting conversation\\ \hline
\end{tabular}
\label{table:actions}
\end{table}

\subsection{Annotation and Analyses}
\begin{figure}[h]
\centering
\includegraphics[width=0.4\textwidth]{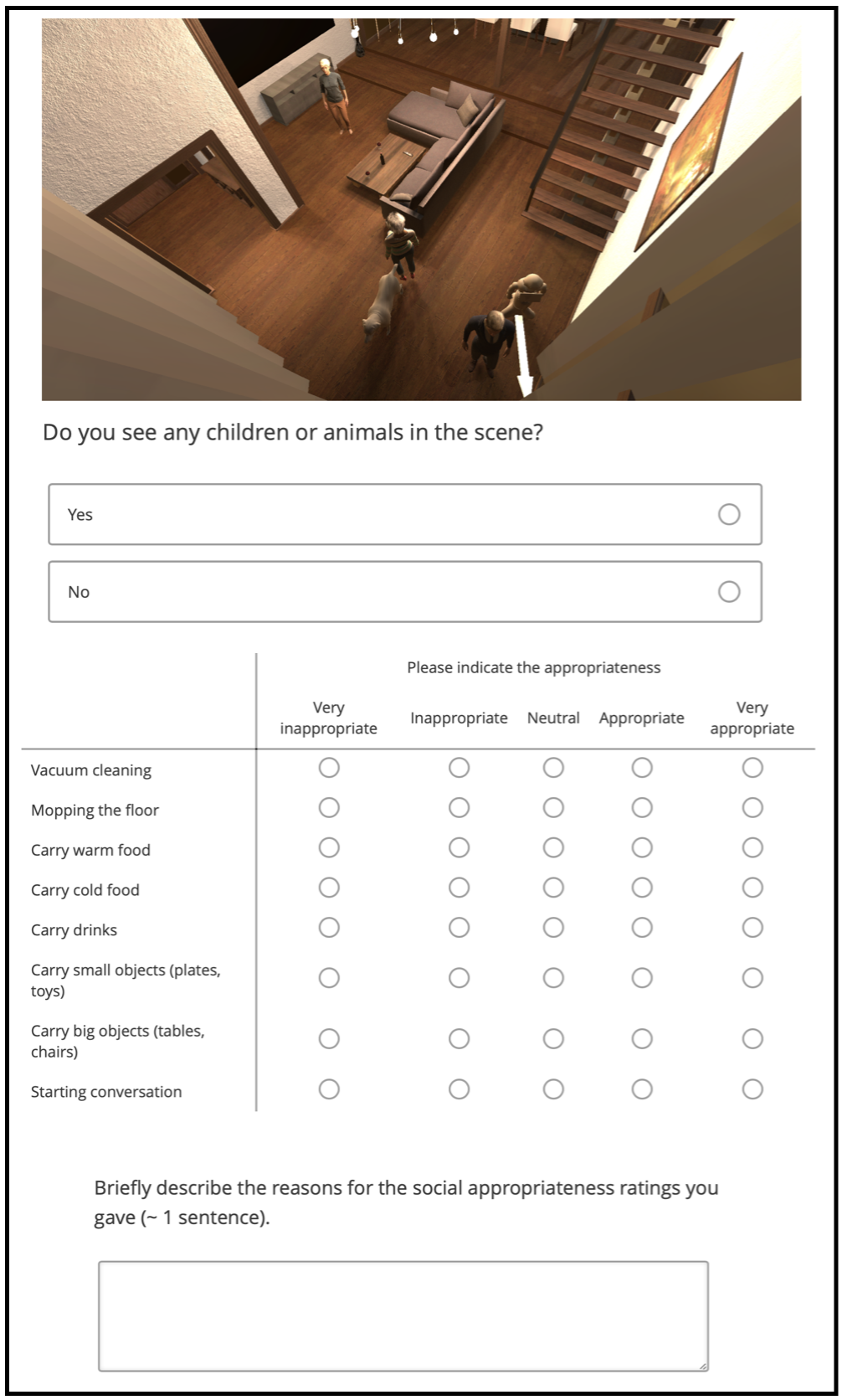}
\caption{The task as shown to the annotators on the crowd-sourcing platform.}
\label{figure:survey}
\end{figure}

\textbf{Data Annotation}. The generated scenes were labelled for social appropriateness of the robot actions in the depicted domestic setting using a crowd-sourcing platform \cite{morris2014crowdsourcing}. The screenshot of what the annotators were presented with is depicted in Figure \ref{figure:survey}. Using this platform, we gathered 15 subjective opinions per scene, on a Likert scale from 1 to 5, where 1 represented ``very inappropriate" and 5 ``very appropriate". 
The annotators constitute a varied group of English speakers. In order to avoid low-quality annotations, participants had to  answer a honeypot question (similarly to \cite{SalamEtAl2017}), that asked them whether there was an animal or child present in the scene (Figure \ref{figure:survey}). They were additionally requested to explain their annotation via free-form sentences in a text box. Once the annotations have been obtained, we first analyze the quality of the annotations and what we can infer from them about the factors affecting social appropriateness of robot actions.

\textbf{Reliability}. We analyzed the reliability of the annotations using Cronbach’s $\alpha$ \cite{bland1997statistics} metric, which tests the reliability of our crowd-sourced data by looking  at  internal  consistency. For the actions-in-circle we obtain $\alpha= 0.885$ and for  actions-along-arrow, $\alpha=0.851$.  According to \cite{alpha_good}, Cronbach’s $\alpha$ values over 0.70 are deemed as a good level of agreement.

\begin{figure}[h]
\centering
\subfigure[]{\includegraphics[width=0.4\textwidth]{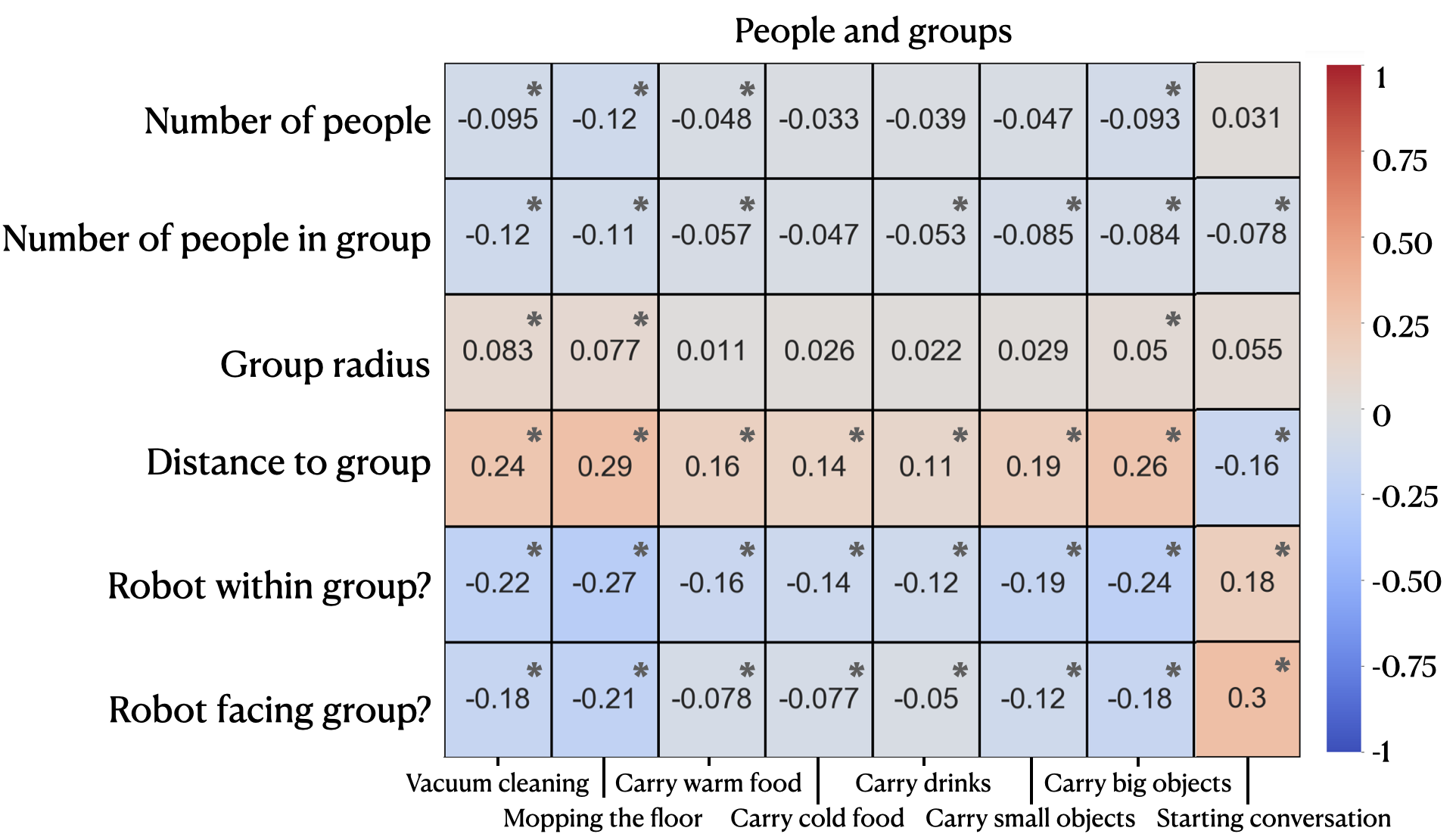}}

\subfigure[]{\includegraphics[width=0.4\textwidth]{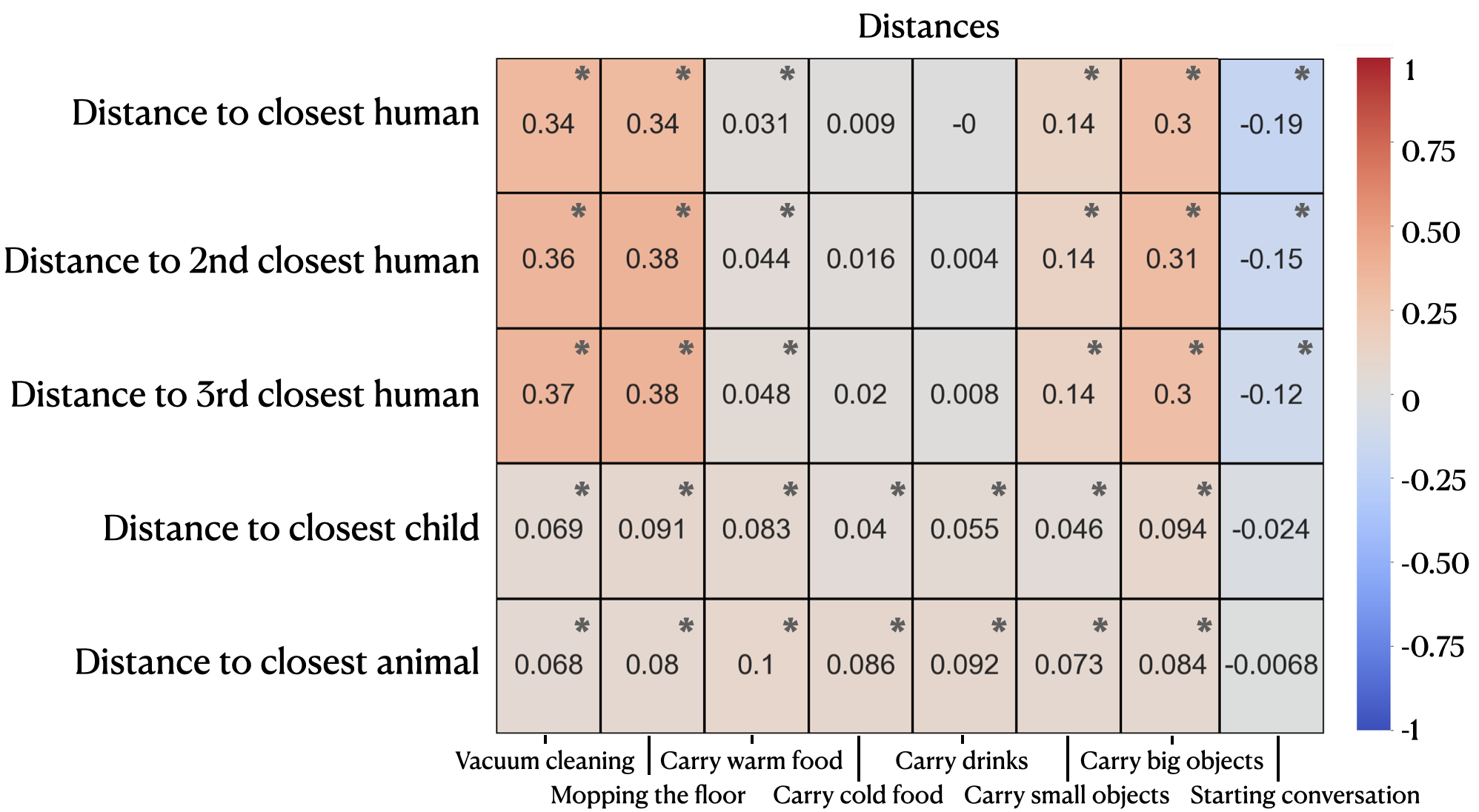}}

\caption{The Pearson correlation \cite{freedman2007statistics} between social appropriateness of actions and (a) group related features, or (b) distance-related features. Values marked with * are statistically significant (p $<$ 0.01).}
\label{fig:correlations}
\end{figure}

\begin{figure}[h]
\centering
\subfigure[]{
\includegraphics[width=0.45\textwidth]{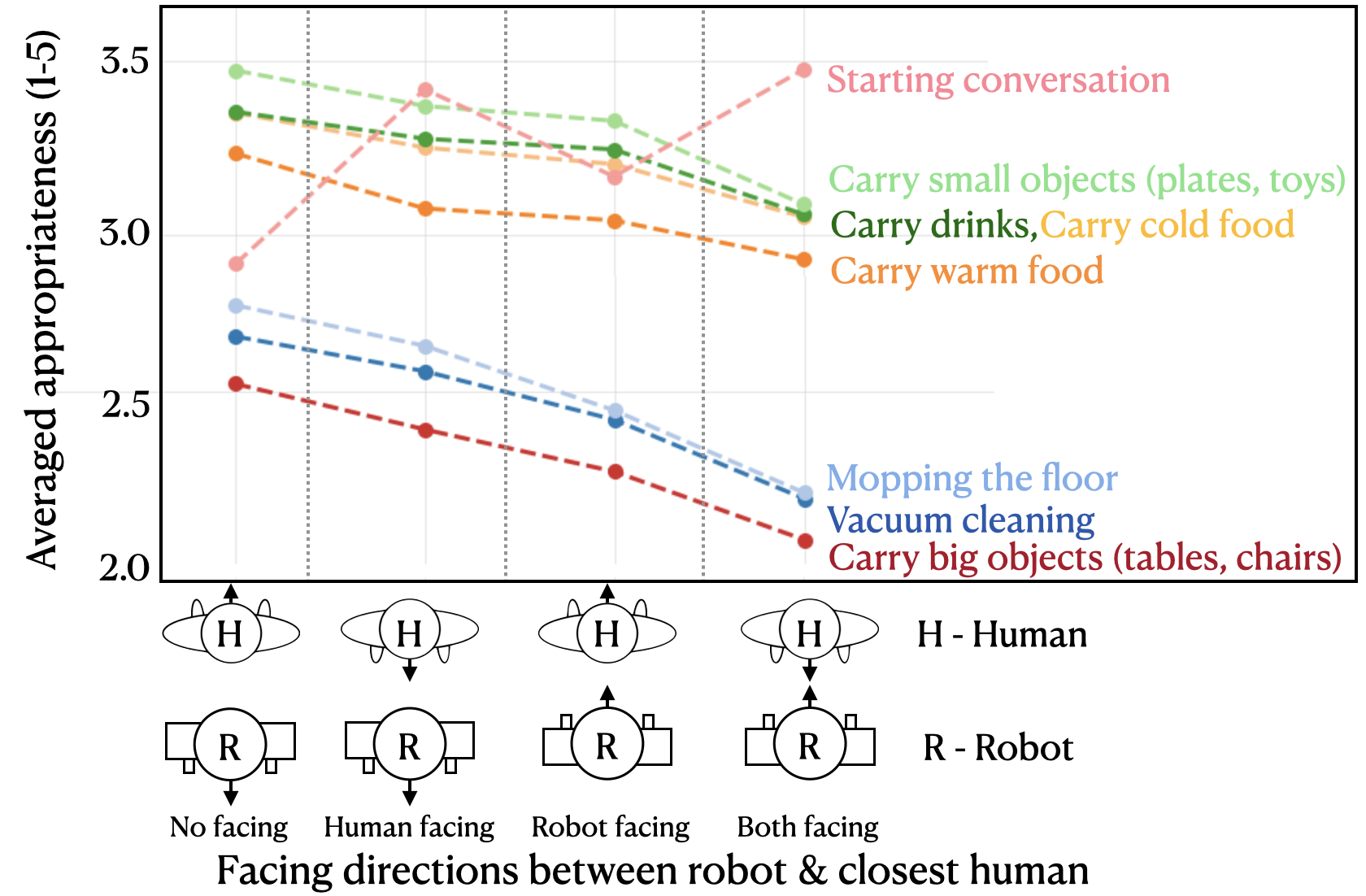}
}
\subfigure[]{
\includegraphics[width=0.45\textwidth]{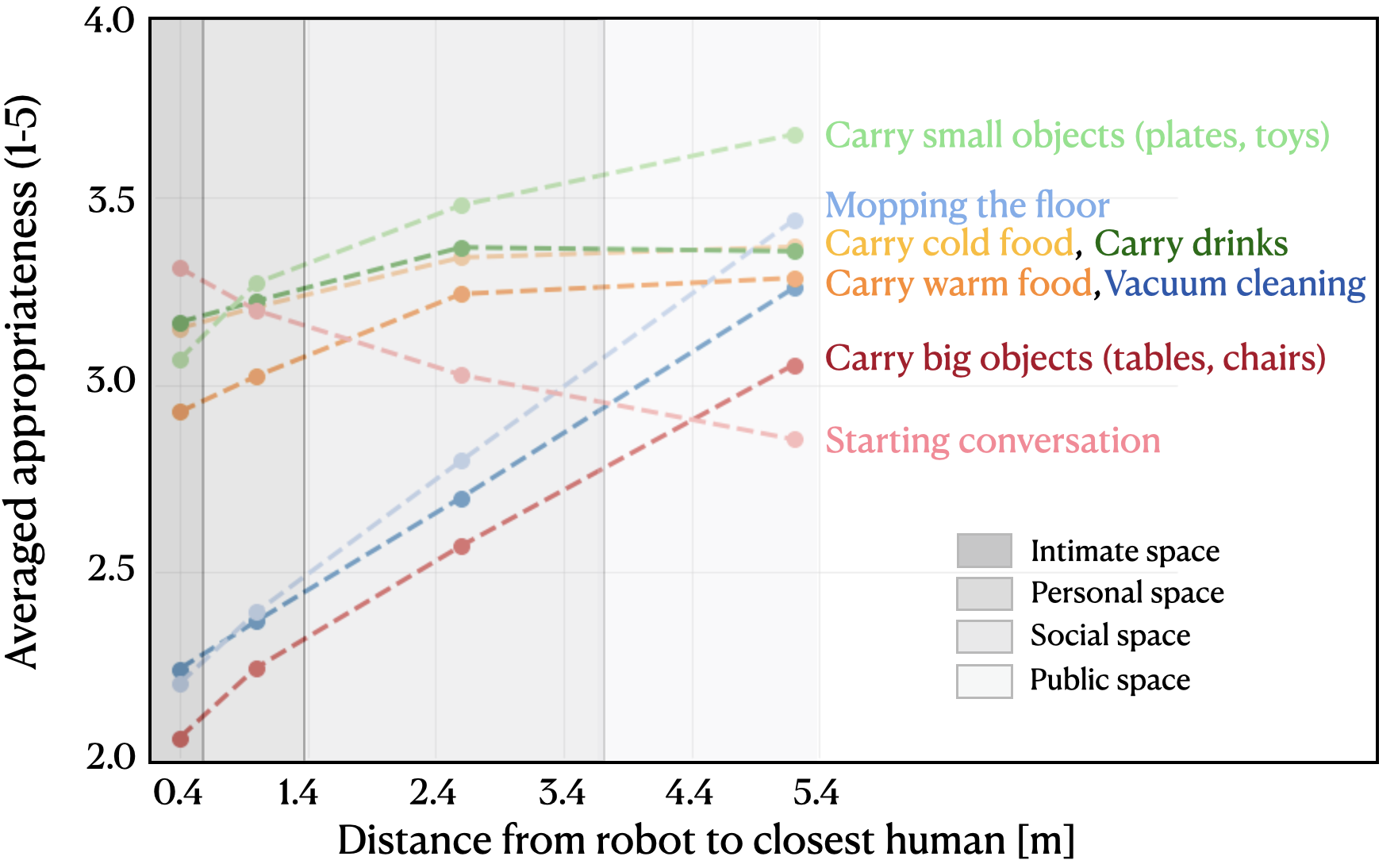}
}
\caption{Average appropriateness of actions with respect to the (a) distance and (b) orientation of the closest person in the environment. \label{fig:dist_and_dir}}
\end{figure}

\textbf{Perceived Social Appropriateness of Actions}. 
We now explore the relation between the various  factors and the social appropriateness of actions. Figure \ref{fig:correlations} provides the Pearson correlation coefficients for group-related and distance-related features separately, where we observe that appropriateness of actions such as ``Vacuum cleaning", ``mopping the floor", ``carry big objects" is affected by the distance to the closest group. Moreover, appropriateness of ``starting a conversation" is strongly affected by whether the robot is facing the group or not. Since distance and orientation are important factors in perceived appropriateness of actions, in Figure \ref{fig:dist_and_dir} we provide an analysis of appropriateness with respect to the distance and the orientation of the closest human. We see that actions that may pose danger to or disturb humans (e.g. carrying big objects, vacuum cleaning) are not perceived as appropriate when the robot is positioned very close to the humans. Instead, for starting a conversation we see an inverse relation, i.e., it is more appropriate to start a conversation when the robot is close to the human. 

\section{CONTINUAL LEARNING}

\subsection{Architecture and Continual Learning Models.}
\begin{figure}[h]
\centering
\includegraphics[width=0.4\textwidth]{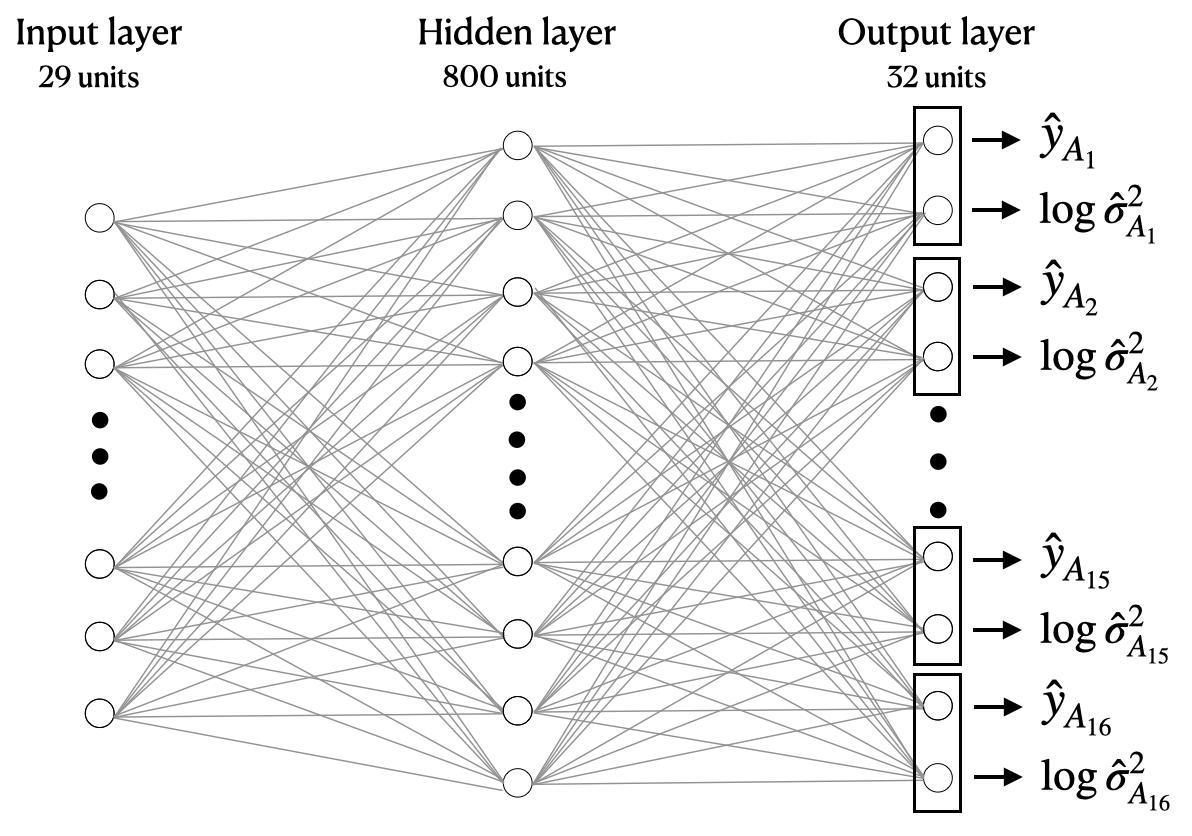}
\caption{Neural network architecture for all three Bayesian models. The models take in the representation of the scene as a 29-dimensional vector (Table \ref{table:feat}) and estimate social appropriateness of each action ($\hat{y}_{A_i}$) in range (1-5) as well as the uncertainty of estimation ($\log \hat{\sigma}_{A_i}$).}
\label{arch}
\end{figure}

We define a two-layer Bayesian Neural Network (BNN), as shown in Fig. \ref{arch}, as the base architecture for estimating appropriateness $\hat{y}_{A_i}$ (in 1-5) as well as its uncertainty $\log \hat{\sigma}_{A_i}$ for each action $A_i$. 

Based on this architecture, we define three continual learning models: 
\begin{itemize}
    \item Baseline (BNN): The model with no continual learning. 
    \item 2-tasks model (BNN-2CL): The model that first learns actions within a circle and then continues learning the actions along an arrow. 
    \item 16-tasks model (BNN-16CL): The model that learns actions (tasks) continually in a sequential manner (inspired by \cite{ChuramaniGunesFG20}).
\end{itemize}

The input to BNN is a 29-dimensional vector that consists of features detailed in Table \ref{table:feat}. These were utilised by the simulation environment when originally generating the MANNERS-DB.

\begin{table}[ht]
\caption{The factors forming the 29-dimensional input to the learning models.}
\centering
\footnotesize
\begin{tabular}{|l|c|c|}  \hline
\emph{Feature} & \emph{Variable type} & \emph{Range}  \\ \hline \hline
Operating within circle & Int & 0 or 1 \\ 
\hline
Radius of action circle & Float & $0.5 \rightarrow 3$ \\ 
\hline
Operating in the direction of an arrow & Int & 0 or 1 \\ 
\hline
Number of humans & Int & $0 \rightarrow 9$ \\ 
\hline
Number of children & Int & $0 \rightarrow 2$ \\ 
\hline
Distance to closet child & Float & $0.4 \rightarrow 6$ \\
\hline
Number of animals & Int & 0 or 1 \\ 
\hline
Distance to animal & Float & $0.4 \rightarrow 6$ \\
\hline
Number of people in a group & Int & $2 \rightarrow 5$ \\ 
\hline
Group radius & Float & $0.50 \rightarrow 1$ \\
\hline
Distance to group & Float & $0 \rightarrow 6$ \\
\hline
Robot within group? & Int & 0 or 1 \\
\hline
Robot facing group? & Int & 0 or 1  \\
\hline
Distance to 3 closest humans & 3 x Float & $0.3 \rightarrow 5$ \\
\hline
Direction robot to 3 closest humans & 3 x Float & $0.0 \rightarrow 360.0$ \\
\hline
Direction closest human to robot & Float & $0.0 \rightarrow 360.0$ \\
\hline
Robot facing 3 closest humans? & 3 x Int & 0 or 1 \\
\hline
3 closest humans facing robot? & 3 x Int & 0 or 1 \\
\hline
Number of people sofa & Int & $0 \rightarrow 2$ \\ 
\hline
Playing music? & Int & 0 or 1 \\
\hline
Total number of agents in scene & Int & $1 \rightarrow 11$ \\ \hline
\end{tabular}
\label{table:feat}
\end{table}

\subsection{Training} 
Unlike conventional neural networks, a BNN models a Normal distribution (with mean \& variance) over parameters, i.e. $\bm{\omega}_i=(\mu_i, \sigma_i)$. Training BNNs is challenging because $p(\bm{\omega}|\bm{X},\bm{Y})$, where $\bm{X},\bm{Y}$ are the inputs and outputs, is intractable. Therefore, we used a backpropagation-compatible approximation, called \textit{Bayes-by-backprop}  \cite{blundell2015weight}, which essentially approximates $p(\bm{\omega}|\bm{X},\bm{Y})$ with a distribution $q_{\bm{\theta}}(\bm{\omega_i})$ whose parameters can be learned.

The loss function minimized for training the models in this fashion is defined (for each action) as follows:
\begin{align}
\mathcal{L}(\bm{\theta}) & = \sum_{i = 1}^{M} 
p_1\log q_{\bm{\theta}}(\bm{\omega_i}) - p_2\log p(\bm{\omega_i})\\
 & + \frac{p_3}{K}\sum_{i = 1}^{K} \frac{1}{2} \hat\sigma^{-2}_i\|y_i -\hat{y}_i\|^2 +\frac{1}{2}\log\hat{\sigma}^2_i,
\end{align}
where the first line controls variational approximation; the second line enforces the correctness of the predictions and estimates their uncertainty; and $p_1= 0.001, p_2= 0.001, p_3= 0.05$ are empirically tuned constants.

When undertaking continual learning, we need to deal with catastrophic forgetting. To prevent this, we employ the uncertainty-guided continual learning strategy in \cite{ebrahimi2019uncertainty}. This method proposes rescaling the learning rate ($\eta$) to calculate a learning rate ($\eta_{\mu_i},\eta_{\sigma_i}$) for each parameter ($\bm{\omega}_i=(\mu_i, \sigma_i)$) according to the current variance $\sigma_i$ of that parameter: $\eta_{\mu_i} \leftarrow \sigma_i \eta$. Following Ebrahimi et al. \cite{ebrahimi2019uncertainty}, we take $\eta_{\sigma_i} = \eta$.

\section{EXPERIMENTS AND RESULTS}

\subsection{Experiments}

\textbf{Implementation Details}. 
We kept all hyperparameters the same for the three models, to allow for a reasonable comparison in performance. Nevertheless, an extensive hyperparameter search was carried out to validate that this did not lead to a substantial drop in performance. 
For training, we used a batch size of 64, 200 epochs per task, an initial global learning rate $\eta$ of 0.06. The learning rate $\eta$ was decayed by a factor of 3 every time the validation loss stopped decreasing for 5 epochs in row, similar to traditional adaptive learning rate methods. Following \cite{ebrahimi2019uncertainty}, we used 10 Monte Carlo samples to approximate  the variational posterior, $q_{\theta}(\omega)$, and the initial mean of the posterior was sampled from a Gaussian centered at 0 with 0.1 in standard deviation. The standard deviation of the weights was initialised as -3. 
Training on each task was done sequentially and the models' weights were saved between tasks. This way, the change in performance, both the ability to predict accurate appropriateness and obtain sensible uncertainty measures, can be investigated with respect to the number of tasks the model has been trained on. It is important to note that the Continual Learning approach adopted here is fundamentally different than many other CL applications. In our work, there is no substantial difference in the input features between data samples of two different tasks. There is, however, difference in terms of labels, as for every new task the model tries to learn to label a new set of actions.

\textbf{Training and Test Sets}. For all three models we split the dataset into training, validation and testing sets. The number of test samples, 100 scenes, are the same for all models, the training and validation set is, however, separated differently to facilitate Continual Learning. The 650 scenes used for training and validation contain 9584 individual labelled samples. The validation part consist of 1000 samples for the BNN model, 400 samples per task (circle and arrow) for the BNN-2CL model and 100 per task (each action) for the BNN-16CL model. This means that the size of the training set for each model is approximately 8500 for the BNN model, 4400 per task for the BNN-2CL model and 500 per task for the BNN-16CL model. It is worth noting that these differences in size of training set affect the comparative results obtained for each model as discussed in the next section.

\subsection{Results}
The prediction results of the three models are presented in Table \ref{tab:prediction_results}. We see that all three models generally estimate the appropriateness level (1-5) with low error (on average, with RMSE values lower than 0.63, for all models). 
Therefore we conclude that the social appropriateness of robot actions can be predicted with a satisfactory level of precision on the MANNERS-DB.

\begin{table}[ht]
\caption{Root-mean-squared error (RMSE) of predictions. \label{tab:prediction_results}}
\centering
\scriptsize
\begin{tabular}{|c|c|c|c|}\hline
\multicolumn{1}{|c|}{\textbf{\textit{Actions}}} & \multicolumn{3}{c|}{\textbf{\textit{RMSE}}}\\ 
\multicolumn{1}{|c|}{\emph{Within a circle}} & \multicolumn{1}{|c|}{\emph{BNN}} & \multicolumn{1}{|c|}{\emph{BNN-2CL}} & \multicolumn{1}{|c|}{\emph{BNN-16CL} } \\ \hline\hline
\multicolumn{1}{|l|}{Vacuum cleaning} & 0.467 & 0.501 & 0.767\\ 
\hline
\multicolumn{1}{|l|}{Mopping the floor} & 0.502 & 0.594 & 0.581\\ 
\hline
\multicolumn{1}{|l|}{Carry warm food} & 0.445 & 0.448 & 0.810 \\ 
\hline
\multicolumn{1}{|l|}{Carry cold food} & 0.420 & 0.403 & 0.561 \\ 
\hline
\multicolumn{1}{|l|}{Carry drinks} & 0.402 & 0.485 & 0.733 \\ 
\hline
\multicolumn{1}{|l|}{Carry small objects} & 0.386 & 0.879 & 0.517 \\ 
\hline
\multicolumn{1}{|l|}{Carry big objects} & 0.497 & 0.520 & 0.665 \\ 
\hline
\multicolumn{1}{|l|}{Cleaning (Picking up stuff)} & 0.192 & 0.479 & 0.491 \\ 
\hline\hline
\emph{In direction of arrow} & \emph{BNN} & \emph{BNN-2CL} & \emph{BNN-16CL}  \\ \hline\hline
\multicolumn{1}{|l|}{Vacuum cleaning} & 0.555 & 0.591 & 0.750\\ 
\hline
\multicolumn{1}{|l|}{Mopping the floor} & 0.542 & 0.602 & 0.664 \\ 
\hline
\multicolumn{1}{|l|}{Carry warm food} & 0.468 & 0.489 & 0.678 \\ 
\hline
\multicolumn{1}{|l|}{Carry cold food} & 0.477 & 0.495 & 0.526 \\ 
\hline
\multicolumn{1}{|l|}{Carry drinks} & 0.451 & 0.465 & 0.586 \\ 
\hline
\multicolumn{1}{|l|}{Carry small objects} & 0.431 & 0.464 & 0.548 \\ 
\hline
\multicolumn{1}{|l|}{Carry big objects} & 0.498 & 0.535 & 0.594 \\ 
\hline
\multicolumn{1}{|l|}{Starting a conversation} & 0.539 & 0.523 & 0.678\\\hline\hline
\multicolumn{1}{|l|}{\textbf{Mean over all actions}} & \textbf{0.480} & \textbf{0.530} & \textbf{0.630}\\
\hline
\end{tabular}
\end{table}

\begin{figure}[ht]
\subfigure[Actions 1-8 (within circle)]{
  \includegraphics[width=0.4\textwidth]{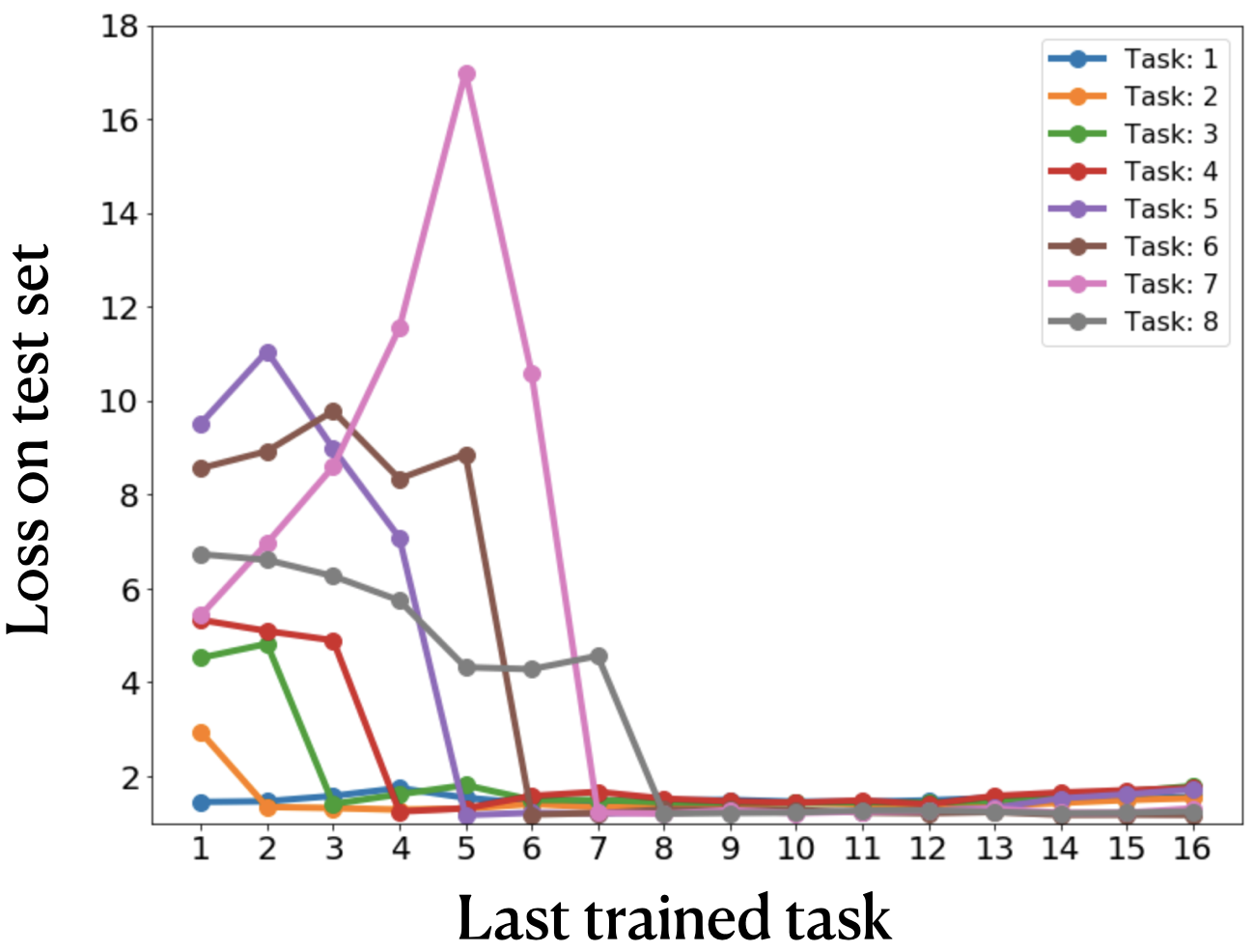}
 }
\subfigure[Actions 9-16 (along arrow)]{\includegraphics[width=0.4\textwidth]{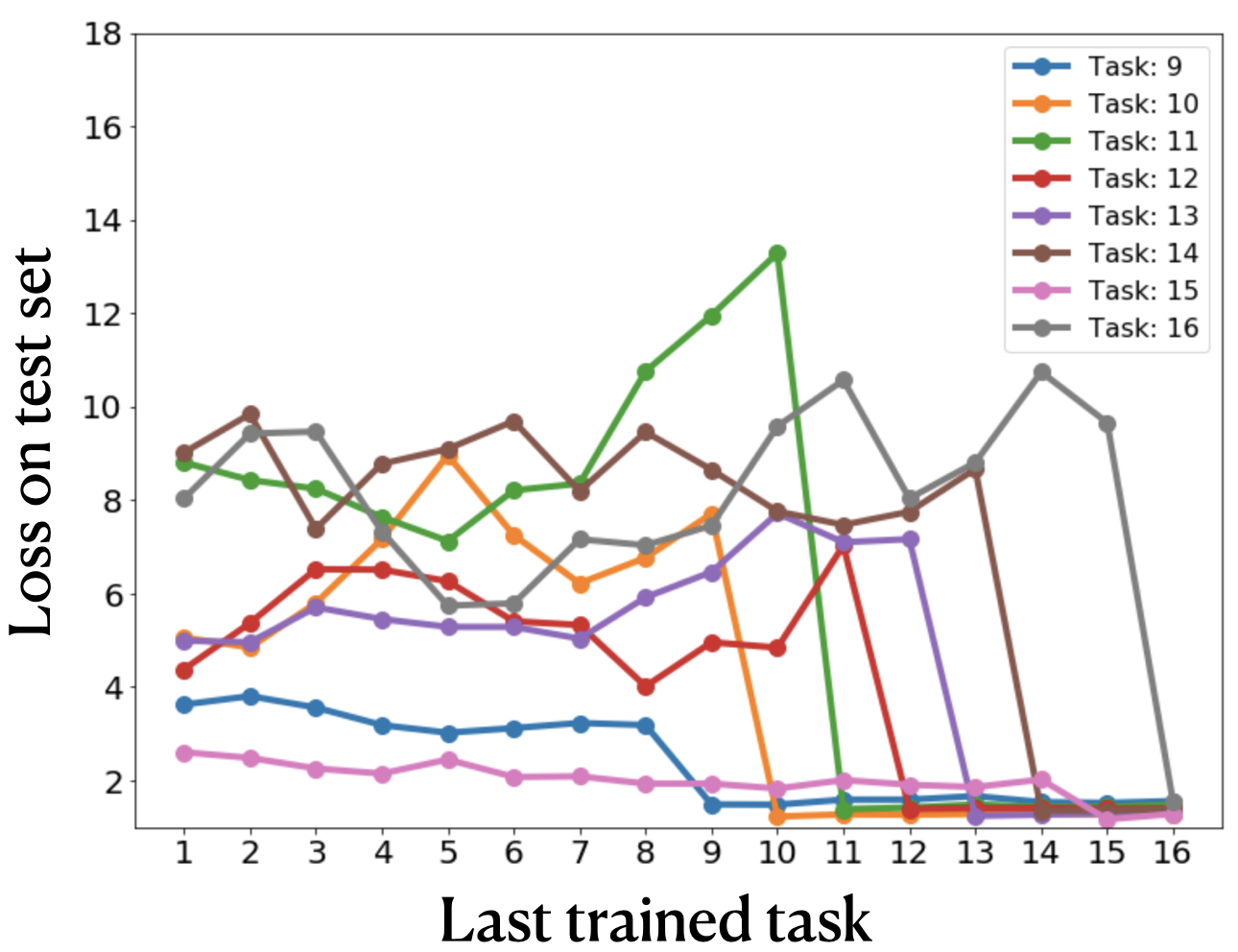}
}
\caption{Per action performance on test the full set at different stages of continual learning.}
\label{cl_loss}
\end{figure}

We provide an analysis of one of the continual learning model's performance (BNN-16CL) in Figure \ref{cl_loss}. The figure shows that there is substantial difference in performance before and after training on a task/action. It also indicates clearly that before a task is trained on, its performance is affected by the training on other tasks. Looking at Figure \ref{cl_loss}(a) and task 7, we observe a good example of this, where the loss is increasing as the model is getting trained on other tasks, before dropping after being trained on the specific task at hand. Looking at Figure \ref{cl_loss}, we confirm that the loss on the test data for a specific task drops as the model gets trained on that specific task and thereafter, stays reasonably low and unaffected by the follow-up training process(es). This suggests that the model is able to handle catastrophic forgetting well.

\section{Conclusion}

In this work, we studied the problem of social appropriateness of domestic robot actions -- to the best of our knowledge, we are the first to do so. For this end, we first introduced a dataset with social appropriateness annotations of robot actions in static scenes generated using a simulation environment. The subjective appropriateness annotations were obtained using a crowd-sourcing platform. 

Our analysis of the annotations revealed that human annotators do perceive appropriateness of robot actions differently based on social context. We identified, for example, starting a conversation is perceived more appropriate if the robot is close to and facing the human. We then formulated learning of social appropriateness of actions as a lifelong learning problem. We implemented three Bayesian Neural Networks, two of which employed continual learning. Our experiments demonstrated that all models provided a reasonable level of prediction performance and the continual learning models were able to cope well with catastrophic forgetting. 

\begin{figure}[h]
\centering
\includegraphics[width=0.49\textwidth]{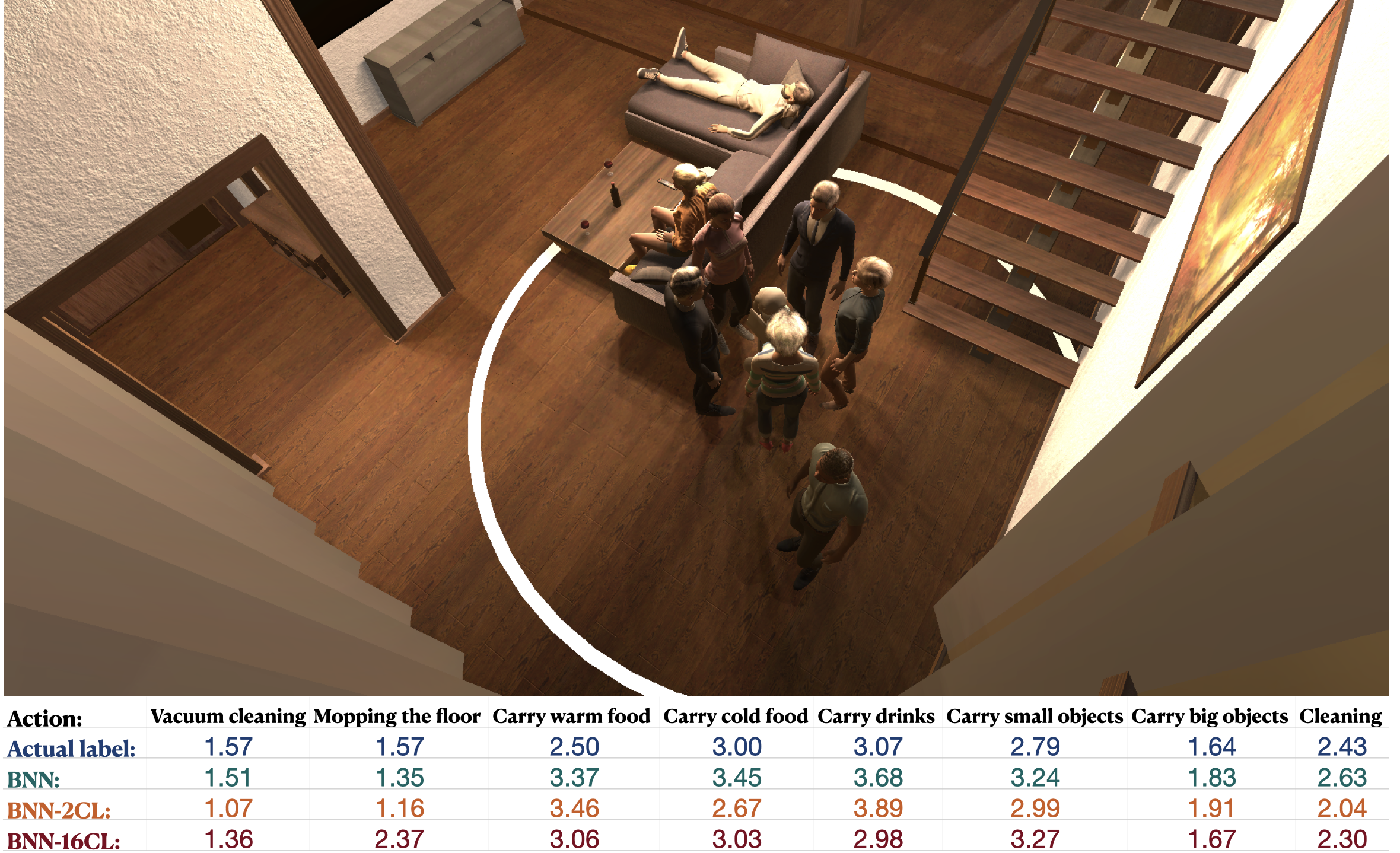}
\caption{Predictions for test scene with people, group and music. The robot is at the center of the circle.}
\label{figure:visual_result}
\end{figure}

In Figure \ref{figure:visual_result}, we provide predictions of the learning models and compare them with the annotations provided by annotators. We observe that the estimated appropriateness levels deviate at most by 1 unit and generally follow the appropriateness-inappropriateness trend of actions; i.e. they are low when the annotations are low, and vice versa. Therefore we conclude that, with the generation of the MANNERS-DB and the proposed learning models, this work takes robots one step closer to a human-like understanding of (social) appropriateness of actions, with respect to the social context they operate in. 

\bibliographystyle{IEEEtran}
\bibliography{references.bib}

\end{document}